\documentclass[journal]{IEEEtran}
\ifCLASSINFOpdf
\else
\fi
\usepackage{url}
\usepackage{booktabs}
\usepackage{multirow}

\usepackage{mathrsfs}
\usepackage{verbatim}
\usepackage{amsmath,amssymb,amsfonts}
\usepackage{lineno}
\usepackage{stfloats}
\usepackage{float}
\usepackage{graphicx}
\usepackage{subfigure}
\usepackage{epstopdf}
\usepackage{textcomp}
\usepackage{bm}
\usepackage{color} 
\usepackage{hyperref}
\usepackage[linesnumbered,ruled,lined,boxed]{algorithm2e}

\usepackage{adjustbox}
\usepackage[numbers,sort&compress]{natbib}
\usepackage{threeparttable}
\begin{document}
\newcommand{\etal}{\textit{et al}. }
\newcommand{\ie}{\textit{i}.\textit{e}., }
\newcommand{\eg}{\textit{e}.\textit{g}. }
%
\title{SCLIFD:Supervised Contrastive Knowledge Distillation for Incremental Fault Diagnosis under Limited Fault Data}
%
%
%

\author{Peng Peng$^\dagger$,~\IEEEmembership{Member,~IEEE},
        Hanrong Zhang$^\dagger$, Mengxuan Li, Gongzhuang Peng, Hongwei Wang$^\ast$,~\IEEEmembership{Member,~IEEE}, and
        Weiming Shen$^\ast$,~\IEEEmembership{Fellow,~IEEE}

\thanks{$^\dagger$Equal contribution.}
\thanks{$^\ast$Corresponding author.}
   
\thanks{This work was supported by the National Natural Science Foundation of
China under Grant 62276230. (Corresponding authors: Hongwei Wang and
Weiming Shen.)}

\thanks{
Peng Peng, Hanrong Zhang, and Hongwei Wang are with Zhejiang University and the University of Illinois Urbana–Champaign Joint Institute, Haining, 314400, China. (E-mail: pengpeng@intl.zju.edu.cn, hanrong.22@intl.zju.edu.cn and hongweiwang@intl.zju.edu.cn). Mengxuan Li is with the College of Computer Science and Technology in Zhejiang University, Hangzhou, 310013, China. (E-mail: mengxuanli@intl.zju.edu.cn).
}%
\thanks{Gongzhuang Peng is
with the National Engineering Research Center for Advanced Rolling
and Intelligent Manufacturing, University of Science and Technology
Beijing, Beijing, 100083, China (E-mail: gzpeng@ustb.edu.cn).}
\thanks{This work has been submitted to the IEEE Transactions on Neural Networks and Learning Systems
for possible publication. Copyright may be transferred without notice, after
which this version may no longer be accessible.}

}

%
%

\markboth{IEEE Transactions on Neural Networks and Learning Systems}%
{Shell \MakeLowercase{\textit{et al.}}: Enhance Fault Diagnosis Performance By Neural Network Ari}
%



\maketitle

\begin{abstract}
Intelligent fault diagnosis has made extraordinary advancements currently. Nonetheless, few works tackle class-incremental learning for fault diagnosis under limited fault data, \textit{i}.\textit{e}., imbalanced and long-tailed fault diagnosis, which  brings about various notable challenges. Initially, it is difficult to extract discriminative features from limited fault data. Moreover, a well-trained model must be retrained from scratch to classify the samples from new classes, thus causing a high computational burden and time consumption. Furthermore, the model may suffer from catastrophic forgetting when trained incrementally. Finally, the model decision is biased toward the new classes due to the class imbalance. The problems can consequently lead to performance degradation of fault diagnosis models. Accordingly, we introduce a supervised contrastive knowledge distillation for incremental fault diagnosis under limited fault data (SCLIFD) framework to address these issues, which extends the classical incremental classifier and representation learning (iCaRL) framework from three perspectives. Primarily, we adopt supervised contrastive knowledge distillation (KD) to enhance its representation learning capability under limited fault data. Moreover, we propose a novel prioritized exemplar selection method adaptive herding (AdaHerding) to restrict the increase of the computational burden, which is also combined with KD to alleviate catastrophic forgetting. Additionally, we adopt the cosine classifier to mitigate the adverse impact of class imbalance. We conduct extensive experiments on simulated and real-world industrial processes under different imbalance ratios. Experimental results show that our SCLIFD outperforms the existing methods by a large margin.
\end{abstract}

%
\IEEEpeerreviewmaketitle

\section{Introduction}
\label{intro}
Frequent occurrences of industrial failures may result in undesirable downtime and productivity loss. Therefore fault diagnosis, which can identify the failure reasons and diminish potential hazards, is crucial to the performance enhancement of industrial systems \cite{Zou_Xia_Li_2018}. Notably, data-driving fault diagnosis methods have become increasingly renowned in the last two decades \cite{Chen_Chen_Fan_Peng_Yang_2022, Wang_Liu_Lin_Chen_Li_Hu_Chen_2021}. Nevertheless, most of them  require sufficient training data for reliable modeling performance \cite{Peng_Lu_Tao_Ma_Zhang_Wang_Zhang_2022}.

Unfortunately, fault data is commonly limited compared with normal data,
because engineering equipment usually operates under normal circumstances and probabilities of fault vary in different working environments. Moreover, fault simulation experiments are expensive and inevitably have a certain difference from the real industrial environment. These possible reasons consequently bring about the class imbalance and long-tailed distribution between different conditions \cite{Chen_Chen_Feng_Liu_Zhang_Zhang_Xiao_2022}. They usually deteriorate the performance of the model, because the model tends to pay more attention to the normal class and hence ignore the fault classes or tail classes. On account of this, abundant research has been conducted to tackle the problem and fruitful achievements have been produced currently. 

Data resampling, cost-sensitive learning, and information augmentation are three popular and effective methods to address the class imbalance and long tail problem \cite{Chen_Chen_Feng_Liu_Zhang_Zhang_Xiao_2022}. Concretely, data resampling alleviates the class imbalance problem by reconstructing a class-balanced dataset \cite{Khoshgoftaar_Gao_2009a}. 
Cost-sensitive learning intends to make amends for the impact of class imbalance throughout the training process by altering the weights of various classes' influence on the target function \cite{Liu_Li_Zio_2017}. 
Information augmentation-based approaches, mainly consisting of data generation and transfer learning, target to resolve the imbalance problem by utilizing auxiliary data or diagnosing experience from other datasets. Concretely, data generation adopts generative models, such as generative adversarial network \cite{Goodfellow_Pouget-Abadie_Mirza_Xu_Warde-Farley_Ozair_Courville_Bengio_2020} and variational auto-encoders \cite{Li_Jiang_Liu_Zhang_Xu_2021}, to create additional samples to increase the dataset via simulating the original distribution of the source data. 
Transfer learning seeks to transfer information, such as data and features, from the source domain to the target domain in order to improve model performance \cite{Zhiyi_Haidong_Lin_Junsheng_Yu_2020}.
\citet{chu2020feature} utilizes the class activation maps to decouple features into class-specific features and common features, which are utilized to augment tail classes by combining the two kinds of features.
 

Apart from the problem that the amount and kind of fault samples available are very inadequate in the real-world industrial fault diagnosis, additional fault samples of new classes, \textit{i}.\textit{e}. incremental samples can be obtained continuously with the progress of the industrial process. Nonetheless, the methods mentioned above only focus on addressing the class imbalance problem but totally disregard the adaptation ability of the models to recognize new fault samples encountered. Consequently, if the fault diagnosis methods mentioned above are directly used in the incremental scenarios,
each time new class samples emerge they have to abandon all prior efforts and retrain the well-trained model, leading to a high computational cost and time expense. What's more, as incremental learning progresses and new types of data are used to train the model, its performance constantly deteriorates because the new knowledge can overwrite the old knowledge learned previously. The phenomenon is called \textit{catastrophic forgetting} \cite{McCloskey_Cohen_1989}. 
As a result, effective incremental learning methods that can mitigate catastrophic forgetting are of great essence for the fault diagnosis field.

Currently, increasingly more methods based on incremental learning in the fault diagnosis field have been paid much attention to by researchers.
\citet{Picot_Maussion_2018} proposes an automatic classification technique for detecting bearing faults based on PCA continuously updated whenever a new class is recognized.
\citet{Hell_Pestana_Soares_Goliatt_2022} proposes a Self-Organizing Fuzzy classifier based on a nonparametric fuzzy-based classifier designed to handle large-scale, streaming, and situations changing constantly. 
 \citet{Ren_Liu_Wang_Zhang_2022} proposes a heterogeneous sample enhancement network (HSELL-Net) adopting lifelong learning, which employs fault samples from other operation conditions to enhance small samples by domain adaptation.
Nevertheless, either the existing methods are limited in the actual industrial small sample scenarios, or they are required to be trained by the fault samples under other operating conditions.

\begin{figure}[t]
    \centering
    \includegraphics[width=0.48\textwidth]{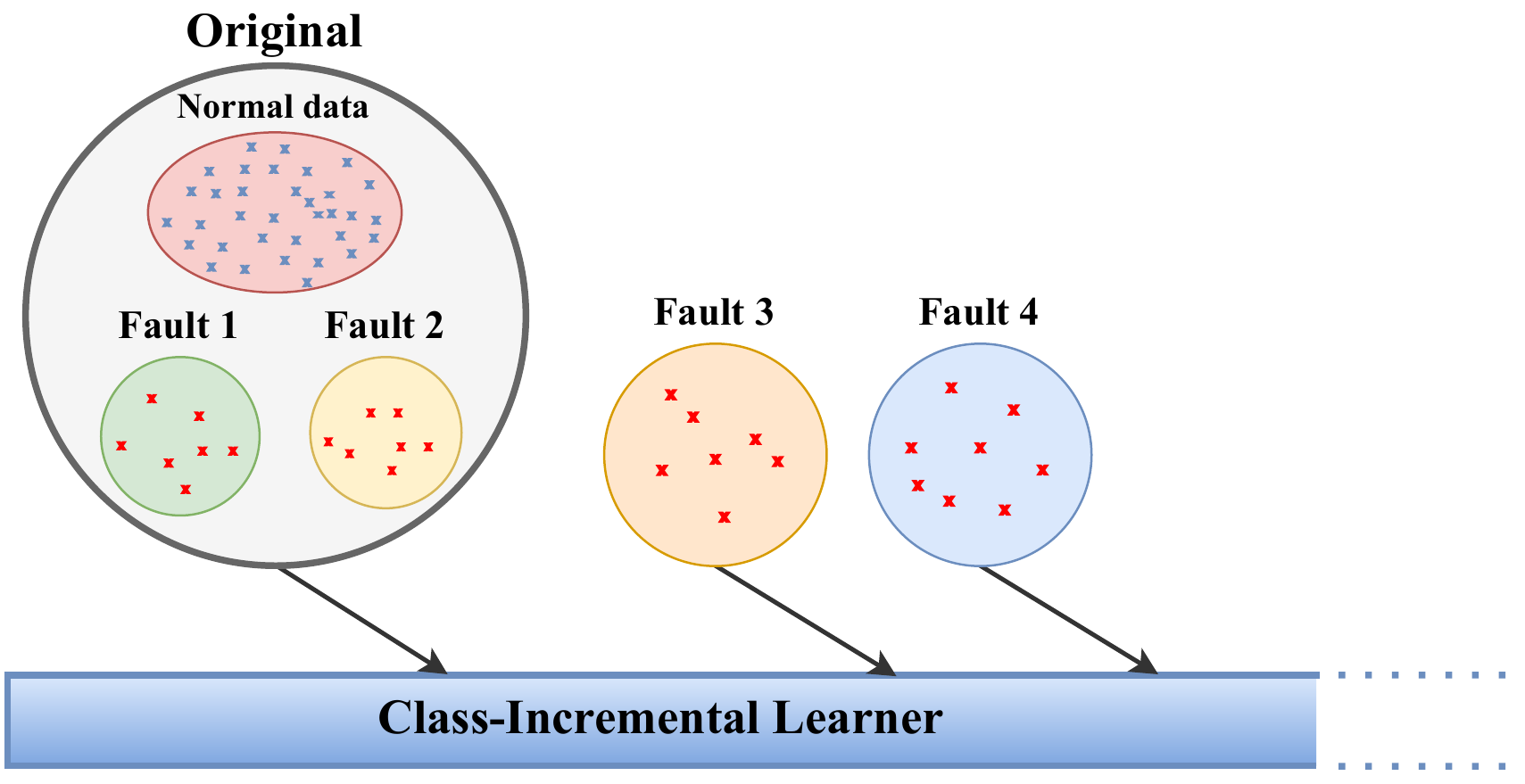}
    \caption{Class incremental learning for fault diagnosis under limited fault data}
    \label{incremental}
\end{figure}

In summary, it is crucial and of great necessity to develop a method that tackles class incremental learning under inadequate data in the fault diagnosis area, as illustrated in Fig. \ref{incremental}. 
The contributions of the paper are summarized as follows:
\begin{enumerate}
    \item We propose the supervised contrastive knowledge distillation for incremental fault diagnosis under limited fault data (SCLIFD) framework based on the classical incremental classifier and representation learning (iCaRL) framework, to address the issues existing in the fault diagnosis models continuously encountering new fault classes with inadequate samples. 
    \item We utilize supervised contrastive learning (SCL) to boost the performance of the feature extractor network, due to the difficulty of learning discriminative features from limited fault samples. What's more, in order to mitigate catastrophic forgetting, we distill the contrast representation of old classes from the feature extractor network in the last incremental session to the network in the current session. 
    \item We propose a novel prioritized exemplar selection method called adaptive herding (AdaHerding) to make better use of prototype rehearsal. It adaptively reserves the most challenging exemplars of each class into a memory buffer for better subsequent training of SCLIFD at each incremental session.
    \item We adopt the cosine classifier to make classifications, which utilizes cosine normalization to alleviate the model decision bias towards new classes arising from the class imbalance.
    \item We evaluate the effectiveness of the proposed method for fault diagnosis on a simulated Tennessee Eastman Process dataset and a practical dataset Multiphase Flow Facility dataset under different imbalance ratios. The results demonstrate that the proposed model achieves the best performance compared with other state-of-the-art methods.
\end{enumerate}

The rest of the paper is organized as follows. In Section \ref{background}, the motivation of the method is highlighted, and the preliminary theories of this work, such as SCL and herding are introduced. We elaborated on the proposed method in Section \ref{method}. The experiments conducted on the TEP and MFF are explained in Section \ref{Experiment}. Section \ref{conclusion} concludes the paper.

\section{Background Theory and Motivation}
\label{background}

\subsection{Motivation}


As we described in Section \ref{intro}, it is vital to address the imbalanced and long-tailed fault diagnosis in the class incremental learning setting, but few works tackle the circumstances well. Accordingly, we consider extending the classical iCaRL framework, which has been proven to be effective in many multi-class incremental learning scenarios, to solve the problem \cite{Tao_Hong_Chang_Dong_Wei_Gong_2020, Castro_2018_ECCV, Wu_2019_CVPR}. 
Intuitively, iCaRL consists of the following three main components:
\begin{itemize}
    \item representation learning based on knowledge distillation
    \item prioritized exemplar selection adopting herding \cite{welling2009herding}
    \item nearest-mean-of-exemplars (NME) classifier
\end{itemize}
Nevertheless, the three components didn't consider the circumstances of class incremental fault diagnosis under limited fault data, leading to poor performance of iCaRL under the circumstances.

Therefore, we consider improving iCaRL in its three main components.
Primarily, iCaRL utilizes cross-entropy loss as the classification loss. However, as we work on the problem, we find that its representation learning capability is insufficient to extract discriminative features from limited data. Therefore, we consider adopting a powerful representation learning method SCL to boost its representation learning performance. 
Furthermore, iCaRL utilizes response knowledge distillation as distillation loss, which mimics the softened softmax output of the teacher network, but it does not effectively prevent the feature extraction network from catastrophic forgetting. To better prohibit the feature network from catastrophic forgetting, we choose to distill the feature representation of the previous feature extractor to the current one.

Moreover, in order to rehearse the exemplars in previous classes, iCaRL adopts herding to construct a memory buffer consisting of the most representative exemplars of each previous class. 
Namely, not only is the training data of the novel classes encountered currently utilized, but also the partial exemplars from previous classes are reserved and used together to update the feature extraction network at each incremental session.
However, the exemplars selected by herding may be easy to be classified, while it is the samples challenging to learn that confuse the model. It is often the case that a model misclassifies the challenging samples to other classes.
The similar thought is also widely used in other scenarios, such as active learning, which selects the samples that the model considers the most difficult to distinguish by some selection strategies for labeling \cite{settles2009active}.
As a result, we consider paying more attention to the samples most difficult to learn when constructing the memory buffer, thus leading to the design of AdaHerding, which adaptively selects the most challenging samples 
according to the number of samples from other classes around them in the embedding space. And the knowledge concerning the samples easy to classify are reserved in the distillation stage.

Finally, there exists a challenge in the imbalance between the old classes observed in earlier stages and the new classes shown in the present stage. Specifically, the model sees few samples from old classes but relatively more samples of novel classes at each incremental session. 
Consequently, the emphasis of the model training is biased toward the newly encountered classes, hence giving rise to classification favoring novel classes, which is a negative impact on the prediction stage. Therefore, in order to handle the problem, we suggest adopting the cosine classifier widely used to alleviate class imbalance to make classification, which utilizes cosine normalization to impose balanced magnitudes across both old and novel classes.  \cite{Hu_Wu_Jian_Chen_Yu_2021, Xu_Yu_Essaf_2022, Koto_2014}.

\subsection{Supervised Contrastive Learning}
\label{SCL}
In this subsection, we will introduce the theory of self-supervised contrastive learning and SCL. The general idea of self-supervised contrastive learning is to make each anchor closer to a positive sample and make the anchor distant from other negative samples in the embedding space \cite{Khosla_Teterwak_Wang_Sarna_Tian_Isola_Maschinot_Liu_Krishnan_2020}.

Specifically, assume a set of $N$ sample/label pairs randomly sampled are denoted as $\left\{\boldsymbol{x}_{k}, \boldsymbol{y}_{k}\right\}_{k=1 \ldots N}$. 
Its augmented batch comprising $2N$ pairs $\left\{\tilde{\boldsymbol{x}}_{i}, \tilde{\boldsymbol{y}}_{i}\right\}_{i=1 \ldots 2 N}$ of the same label with the source sample are generated and used for training. 
The two random augmentations of $\boldsymbol{x}_{k}(k=1 \ldots N)$ are denoted as $\tilde{\boldsymbol{x}}_{2 k}$ and $\tilde{\boldsymbol{x}}_{2 k-1}$, of which arbitrary one is called the anchor and the other one is called the positive. Their index is denoted as $i \in I \equiv\{1 \ldots 2 N\}$ and $j(i)$ respectively. The $2N-2$ samples remained in $\left\{\tilde{\boldsymbol{x}}_{i}, \tilde{\boldsymbol{y}}_{i}\right\}_{i=1 \ldots 2 N}$ are called the negatives.

The self-supervised contrastive loss is defined in the following form:
\begin{equation}
\label{ssl}
    \mathcal{L}^{\text {self }} = \sum_{i \in I} \mathcal{L}_{i}^{\text {self }} = -\sum_{i \in I} \log \frac{\exp \left(\boldsymbol{z}_{i} \cdot \boldsymbol{z}_{j(i)} / \tau\right)}{\sum_{a \in A(i)} \exp \left(\boldsymbol{z}_{i} \cdot \boldsymbol{z}_{a} / \tau\right)}
\end{equation}
Here, $\boldsymbol{z}_{l}=\varphi(\tilde{\boldsymbol{x}}_{l})$ represents the embedding of a sample $\tilde{\boldsymbol{x}}_{l}$ that is the output of a contrastive network, and the $\cdot$ represents the inner dot product. $A(i) \equiv I \backslash\{i\}$ and $\tau \in \mathcal{R}^{+}$ denotes a scalar temperature parameter.

Due to the incapability of the loss in Equation \ref{ssl} to utilize the label information, it is not suitable for fully-supervised circumstances. Accordingly, the  supervised contrastive loss based on self-supervised contrastive loss is proposed to leverage label information \cite{Khosla_Teterwak_Wang_Sarna_Tian_Isola_Maschinot_Liu_Krishnan_2020}. The samples with the same label for each anchor are considered to be positive. The definition of SCL is shown below:
\begin{align}
\label{scl}
\mathcal{L}_{scl} & = \sum_{i \in I} \frac{-1}{|P(i)|} \sum_{p \in P(i)} \log \frac{\exp \left(\boldsymbol{z}_{i} \cdot \boldsymbol{z}_{p} / \tau\right)}{\sum_{a \in A(i)} \exp \left(\boldsymbol{z}_{i} \cdot \boldsymbol{z}_{a} / \tau\right)}
\end{align}
where $P(i) \equiv\left\{p \in A(i): \tilde{\boldsymbol{y}}_{p}=\tilde{\boldsymbol{y}}_{i}\right\}$ is the set of indices of all positives of the anchor which are different from $i$, and $|P(i)|$ is its cardinality.

\subsection{Prioritized Exemplar Selection Method Herding}
\label{sec:herding}
In order to overcome the problem of catastrophic forgetting, iCaRL adopts the principle of rehearsal. Moreover, as more new classes of data are reserved and thus accumulated for training afterward, the computational and storage requirements also increase and the update speed of the model will considerably decrease. To solve the problems, iCaRL keeps a fixed number of $K$ exemplars for the total storage of both old and new classes. Assume currently $t$ classes have been seen, then iCaRL will store $m=K/t$ (up to arounding) exemplars for each class. Hence it is necessary to keep the $m$ most representative exemplars of each class for more effective model training. 

 Whenever iCaRL encounters novel classes, it adopts the same prioritized construction strategy in herding to
 create a set of representative samples of restricted size $m$, from the new class according to the principle of rehearsal. Thus we will refer to the method as herding for the remainder of the paper. 
 Assume the novel class is a set $X = \left\{x_{1}, \ldots, x_{n}\right\}$ of class $y$, and the current feature function is $\varphi: \mathcal{X} \rightarrow \mathbb{R}^{d}$. Then the current class mean $\mu$ is computed as $\mu = \frac{1}{n} \sum_{x \in X} \varphi(x)$. \textit{Herding} selects and stores exemplars $p_{1}, \ldots, p_{m}$ iteratively until the exemplar set reaches the target number $m$. 
 The selection is computed as follows:
 \begin{equation}
     p_{k} = \underset{x \in X}{\operatorname{argmin}}\parallel \mu-\frac{1}{k}[\varphi(x)+\sum_{j = 1}^{k-1} \varphi\left(p_{j}\right)]\parallel 
 \end{equation}
 where $k$ is iterated from 1 to $m$.
 In each iteration, one more exemplar, which makes the average feature vector of the current exemplar set best approximate that over the whole training samples, is added to the current exemplar set. This consequently causes the exemplar set to be actually a prioritized set. Hence the exemplar earlier in the set is more significant than those in the behind \cite{8100070}. 

\subsection{Cosine Classifier}
In order to predict a label $y^{*}$ for a sample $x \in X\equiv\left\{x_{1}, \ldots, x_{n}\right\}$ of a class $y \in Y\equiv\left\{y_{1}, \ldots, y_{t}\right\}$, where $Y$ is the set of classes observed so far,
first, compute a prototype vector for each $y \in Y$, $\mu_{1}, \ldots, \mu_{t}$. 
Let $\varphi({x})$ represent the embedding of a sample ${x}$.
The average feature vector $\mu_{y}$ of all exemplars of a class $y$ is defined as follows.
\begin{align}
\label{average}
\mu_{y} & = \frac{1}{\left|X\right|} \sum_{x \in X} \varphi(x)
\end{align}

Then the sample ${x}$ should be classified into the class label with the most comparable prototype according to the normalized cosine similarity. It is computed as follows.
\begin{align}
\label{classify}
y^{*} & = \underset{y \in Y}{\operatorname{argmax}}\left(\frac{\exp \left(\left\langle\bar{\mu}_{y}, \bar{\varphi}(x)\right\rangle\right)}{\sum_{y\prime \in Y} \exp \left(\left\langle\bar{\mu}_{y\prime}, \bar{\varphi}(x)\right\rangle\right)}\right)
\end{align}
where $\bar{v}=v /\|v\|_{2}$ represents the $l_2$-normalized vector, and $\left\langle\bar{v}_{1}, \bar{v}_{2}\right\rangle=\bar{v}_{1}^{\mathrm{T}} \bar{v}_{2}$ denotes the cosine similarity between two normalized vectors $\bar{v}_{1}$ and $\bar{v}_{2}$.

\section{Methodology}
\begin{figure*}[t]
    \centering
    \includegraphics[width=\textwidth]{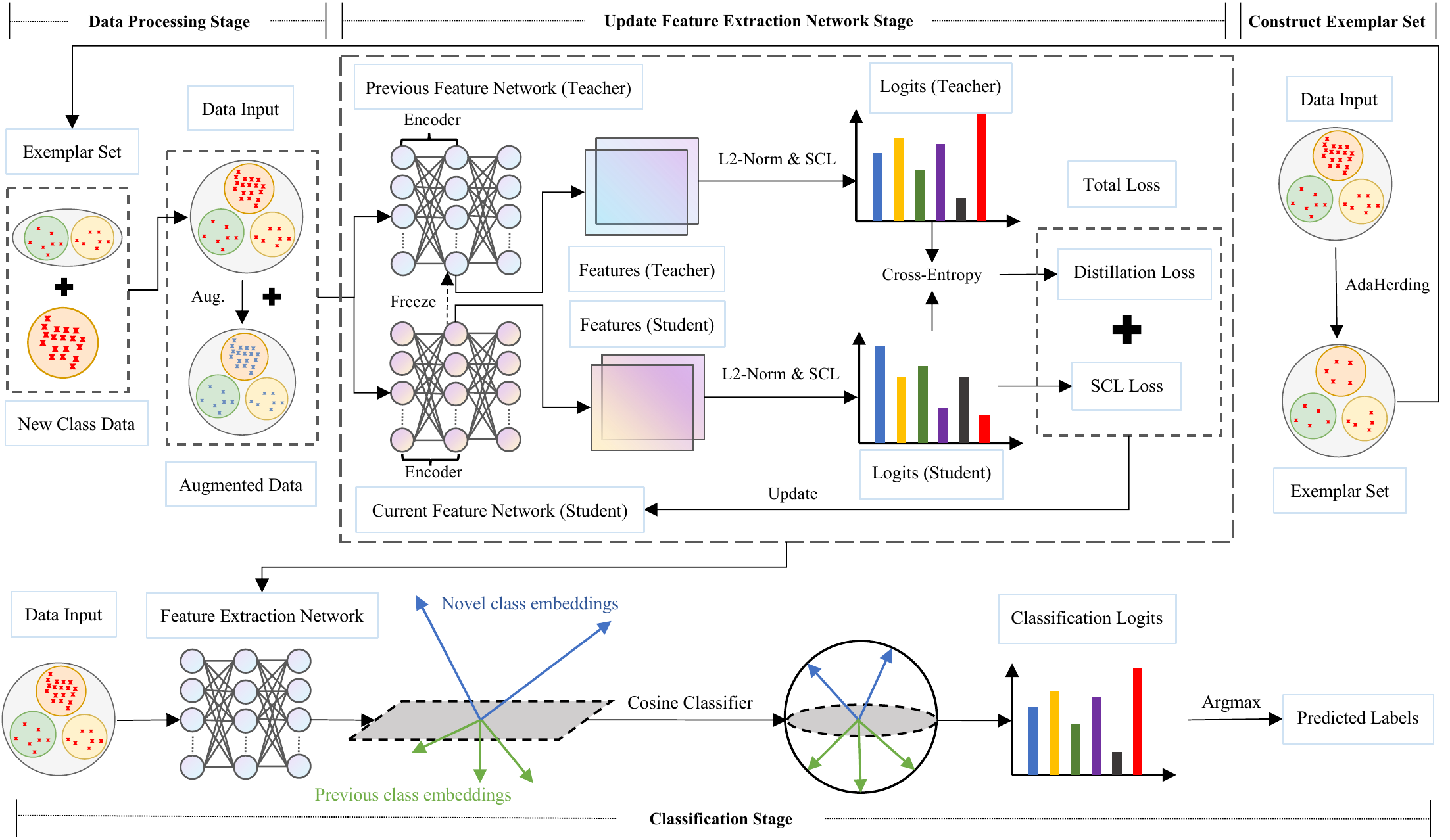}
    \caption{The general framework of the proposed method SCLIFD for fault diagnosis under limited fault data}
    \label{update}
\end{figure*}

\label{method}
As illustrated in Fig. \ref{update}, the general framework of SCLIFD consists of four stages: data processing stage, updating feature extraction network stage, constructing exemplar set stage, and classification stage. In the following subsections, we will elaborate on each stage of SCLIFD and the main flow of class incremental fault diagnosis by SCLIFD in detail.

\subsection{Data Processing}
\label{process}
At each incremental session, SCLIFD encounters samples of new classes. Then the new class data will be combined with the exemplar set constructed by AdaHerding, which will be elaborated on in Section \ref{sec:adaherding}. Specifically, assume the exemplar set consists of $L$ sample/label pairs $\left\{\boldsymbol{x}_{i}, \boldsymbol{y}_{i}\right\}_{i=1 \ldots L}$, given the new class data $\left\{\boldsymbol{x}_{j}, \boldsymbol{y}_{j}\right\}_{j=1 \ldots M}$, the combined data will be $\left\{\boldsymbol{x}_{k}, \boldsymbol{y}_{k}\right\}_{j=1 \ldots L+M}$. Then, the combined data will be augmented. After that, we use the combined data and its augmented data to further train the feature extraction network, so as to adapt the network to the new class data, which will be introduced in the next subsection. 

\subsection{Update Feature Extraction Network}
\label{sec:SCLKD}
As illustrated in Fig. \ref{update}, we utilize supervised contrastive knowledge distillation to enhance the representation learning capability of SCLIFD. We have introduced SCL in detail in section \ref{sec:SCLKD}. In the subsection, we will elaborate on how we combine SCL with knowledge distillation (KD) to train the feature extractor in the stage of representation learning.

Knowledge distillation can effectively enhance the performance of compact models under the supervision of larger pre-trained models. A typical ``teacher-student" KD utilizes a pre-trained larger teacher model to supervise a smaller student model, which is conducive to improving a better representation quality compared to a student model trained from scratch. 
In SCLIFD, we make representation space distillation, which emulates the latent feature space of the teacher. 
In this way, the compact feature representation can be distilled from the backbone of the teacher and for the student to imitate. The teacher model in SCLIFD is the feature extractor network of the last incremental session, and the student model is that of the current session.

Specifically, let $a \in A(i) \equiv I \backslash\{i\}$ 
and $P(\boldsymbol{z}_{i}; \boldsymbol{z}_{a})$ denote the similarity score between $\boldsymbol{z}_{i}$ and $\boldsymbol{z}_{a}$. Then the softmax function is applied with the temperature scaling factor $\tau$ to $P(\boldsymbol{z}_{i}; \boldsymbol{z}_{a})$, resulting in:
\begin{align}
P(\boldsymbol{z}_{i}; \boldsymbol{z}_{a})  & = \frac{\exp \left(\boldsymbol{z}_{i} \cdot \boldsymbol{z}_{a} / \tau\right)}{\sum_{j \in A(i)} \exp \left(\boldsymbol{z}_{i} \cdot \boldsymbol{z}_{j} / \tau\right)}
\end{align}

Finally, we adopt cross-entropy loss to obtain the KD loss, which is defined as follows:
\begin{equation}
\label{dis}
\begin{split}
    \mathcal{L} _ { dis } &=\frac{1}{2 N} \sum_{i \in I} \sum_{a \in A(i)} \mathcal{L}_{i, a}^{d i s} \\
&=-\frac{1}{2 N} \sum_{i \in I} \sum_{a \in A(i)} P(\boldsymbol{z}_{i}; \boldsymbol{z}_{a})^{T} \log \left(P(\boldsymbol{z}_{i}; \boldsymbol{z}_{a})^{S}\right)
\end{split}
\end{equation}
where $P(\boldsymbol{z}_{i}; \boldsymbol{z}_{a})^{T}$ and $P(\boldsymbol{z}_{i}; \boldsymbol{z}_{a})^{S}$ represent the similarity score of the teacher network and student network respectively.

The final total loss is the weighted combination of Equation \ref{scl} SCL loss and Equation \ref{dis} KD loss, which is shown below:
\begin{equation}
    \mathcal{L} = \mathcal{L}_{scl} + \lambda \mathcal{L}_{dis}
\end{equation}
where $\lambda$ is the balancing weight. We set $\lambda=0.5$ in our implmentation.
The final loss will be used to update the current feature network.

\subsection{Construct Exemplar Set by AdaHerding}
\label{sec:adaherding}

SCLIFD adjusts its exemplar set whenever seeing new classes by utilizing the prioritized exemplar selection strategy AdaHerding. 
The same as iCaRL, as is explained in Section \ref{sec:herding}, when 
$t$ classes have been encountered, SCLIFD will store a memory buffer of $K$ exemplars for rehearsal, from which $m = K/t$ exemplars (up to rounding) are allocated for each class. Unlike iCaRL, SCLIFD uses AdaHerding to select the most difficult $m$ exemplars of each class to learn. 

\begin{figure}[t]
    \centering
    \includegraphics[width=0.35\textwidth]{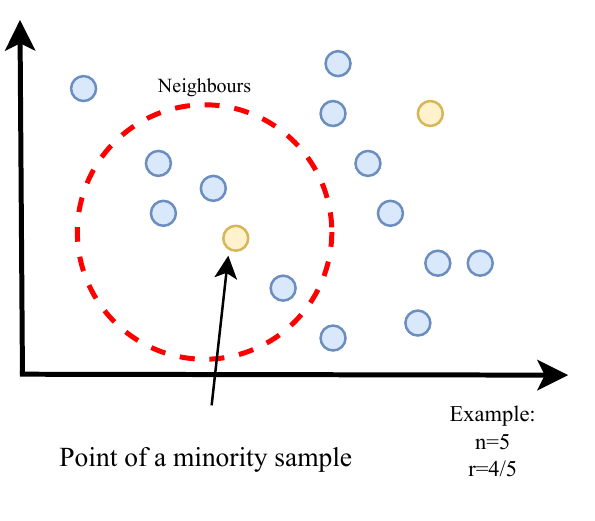}
    \caption{An example of computing the majority ratio in AdaHerding}
    \label{AdaHerding}
\end{figure}

Algorithm \ref{alg:adaherding} explains the exemplar selection process, where the minority class and the majority class indicate the class of the current sample and the other classes respectively. The key idea of AdaHerding is to measure the majority class ratio among the $n$ neighbors of a minority sample. Namely, the more majority class samples around the sample, the more difficult the sample is to learn. Specifically, assume the majority ratio is $r$. Then find the n-nearest neighbors of the sample. Among the neighbors, count the number of samples belonging to the majority class, say $majorityNum$. Then the majority ratio $r=majorityNum/n$. The $r$ value indicates the dominance of the majority class in a specific individual neighborhood. One example is shown in Fig. \ref{AdaHerding}. The blue and yellow points represent the majority and minority class samples respectively. For the 5 neighbors in the red circle, there are 4 majority-class samples. Therefore the $r$ value for the yellow point in the circle is $4/5$.

After calculating the majority ratio of all the samples in the current class, sort the samples according to the majority ratios in descending order. Then choose the first $m$ samples to construct the exemplar set, namely the most challenging $m$ samples with the largest $r$ values.
\begin{algorithm}
    
    \caption{SCLIFD Constructs Exemplar Set by AdaHerding}\label{alg:adaherding}
    \SetKwFunction{isOddNumber}{isOddNumber}

    \SetKwInOut{KwIn}{Input}
    \SetKwInOut{KwOut}{Output}
    \SetKwInOut{KwReq}{Require}
    \KwIn{sample set $X=\left\{x_{1}, \ldots, x_{t}\right\}$ of a class $y$, its labels $Y=\left\{y_{1}, \ldots, y_{t}\right\}$,
    features of all samples previously meet $F=\left\{f_{1}, \ldots, f_{n}\right\}$}
    \KwReq{current feature function $\varphi: \mathcal{X} \rightarrow \mathbb{R}^{d}$, a K-Nearest Neighbour classifier $K$}
    
    \For{x in X}{ 
    $F_x.append(Normalize(\varphi(x)))$ 
    }
    $F.append(F_x)$ 
    
    \tcp{store features of the samples in $y$}
    $K.fit(F,Y)$ \tcp{fit the KNN classifier}
	
    \For{$i=1\ldots t$}{
        $neighbors \leftarrow K.kneighbors(F_x[i],n)$
        
       \tcp{find n neighbors around a sample}
       $majorityNum=CountMajority(neighbors)$

       \tcp{count number of majority class samples among the neighbors}

       $r \leftarrow majorityNum/n$
       
       \tcp{calculate the majority ratio}
       $R[i] \leftarrow r$
       
       \tcp{R is a dictionary mapping sample index to majority ratio}
    }

    $R \leftarrow SortbyValues(R)$

    \tcp{Sort R by majority ratio in descending order}
    $R=\{i_1:r_1,i_2:r_2,\ldots,i_t:r_t\}$

    \tcp{current order of $R$}
    
    $X \leftarrow \{x_{i_1},x_{i_2},\ldots,x_{i_m}\}$ 
    
    \tcp{Keep the first m exemplars}
    \KwOut{$m$ target number of exemplars of $X$}
\end{algorithm}

\begin{table*}[ht]
\centering
\caption{TEP and MFF Dataset Setting in Training and Testing Process}
\label{datasets}
\begin{tabular}{@{}ccccccccc@{}}
\toprule
\multirow{2}{*}{Dateset} & \multirow{2}{*}{Diagnosis Mode} & \multirow{2}{*}{Total Classes} & \multirow{2}{*}{Incremental Sessions} & \multirow{2}{*}{Novel-Class Shot} & \multicolumn{2}{c}{Training set}   & \multicolumn{2}{c}{Testing set}             \\ \cmidrule(l){6-9} 
                         &                                 &                                &                                  &                                   & Normal Size          & Fault  Size & Normal  Size         & Fault  Size          \\ \midrule
\multirow{2}{*}{TEP}     & Imbalanced case 1                     & \multirow{2}{*}{10}            & \multirow{2}{*}{5}               & \multirow{2}{*}{2}                & \multirow{2}{*}{500} & 48          & \multirow{2}{*}{800} & \multirow{2}{*}{800} \\
                         & Imbalanced case 2                     &                                &                                  &                                   &                      & 30          &                      &                      \\ \midrule
\multirow{2}{*}{MFF}     & Imbalanced                      & \multirow{2}{*}{5}             & \multirow{2}{*}{5}               & \multirow{2}{*}{1}                & \multirow{2}{*}{500} & 30          & \multirow{2}{*}{800} & \multirow{2}{*}{800} \\
                         & Long-tailed                     &                                &                                  &                                   &                      & 20          &                      &                      \\ \bottomrule
\end{tabular}
\end{table*}
\subsection{Classification Stage}
\label{sec:classify}
In the classification stage, the feature extraction network initially extracts the data's features. Nevertheless, arising from the class imbalance, the magnitudes of the embeddings of the novel class will be higher than those of the old classes. Accordingly, we apply the cosine classifier to the embeddings above afterward. Due to cosine normalization, the balanced magnitudes of the embeddings will be imposed across both previous and new classes. Then we will compute the average feature vector of each class according to the Equation \ref{average}. Finally, we classify the sample into the most comparable class according to the normalized cosine similarity. The computing method is shown in Equation \ref{classify}.

\subsection{Class Incremental Fault Diagnosis Procedure}
We have detailedly explained the components of SCLIFD above. In the subsection, we will explain the application of the entire process of SCLIFD in class incremental fault diagnosis under limited data, as depicted in Fig. \ref{update}, which is comprised of four steps:
\begin{enumerate}
    \item Process data. With an industrial process continuously going on, the data of new classes emerge accordingly. Then the new data will be processed as explained in Section \ref{process}. Finally, the processed data will be input into the next stage.
    \item Train the feature extraction network. First, the data will be input into the previous and current feature networks, where the latent features output by each network are obtained. After the feature normalization and SCL, the logits of the previous and current feature networks are obtained, based on which the distillation loss is also obtained. Then the SCL loss is obtained based on the latent features output by the current feature network, as explained in Section \ref{sec:SCLKD}. Next, The SCL loss is combined with the distillation loss to constitute the final loss, which is used to update the current feature network. Finally, the updated feature network is frozen to be the previous feature network, which will be utilized in the next incremental session as the teacher network. The updated feature network will also be used in the final fault diagnosis stage.
    \item Construct the exemplar set. This stage constructs an exemplar set of a certain size by the AdaHerding strategy as a memory buffer for prototype rehearsal, as elaborated on in Section \ref{sec:adaherding}. The exemplar set will be used in the initial data process stage with the new class data in the next incremental session.
    \item Online Fault diagnosis. First, the feature extraction network trained previously will be used to learn the representations of the fault samples. Then as explained in Section \ref{sec:classify}, the cosine classifier will be utilized to perform the final online fault diagnosis based on the representations.
\end{enumerate}

\section{Experiment Study}

\label{Experiment}
In this section, we conduct comprehensive experiments on the well-known benchmark dataset Tennessee Eastman Process (TEP) dataset under two imbalanced fault diagnosis situations of different imbalance ratios, which is used as the fundamental validation set.
Furthermore, a practical dataset Multiphase Flow Facility (MFF) case study dataset is used to further evaluate the effectiveness of our proposed model SCLIFD under imbalanced and long-tailed fault diagnosis. 
The classification accuracy of SCLIFD is compared to that of alternative popular methods: Learning without Forgetting (LwF.MC) \cite{Li2018}, Finetuning, iCaRL \cite{8100070}, End-to-End Incremental Learning (EEIL) \cite{Castro_2018_ECCV}, Bias Correction (BiC) Method \cite{Wu_2019_CVPR}. The upper bound of the following experiments is also calculated.  Moreover, ablation experiments are also performed to separate individual components of SCLIFD, which shed light on its working mechanisms. 

\subsection{Evaluation on TEP}

\subsubsection{Dataset Description}

The Tennessee Eastman process (TEP) simulates realistic chemical processes and is especially popular in the fault diagnosis field. The dataset is composed of 52 variables, such as temperature and pressure, which are monitored in real time by sensors. The dataset consists of 20 types of faults, from which we randomly pick nine fault types (type 1, 2, 4, 6, 7, 8, 12, 14, and 18) and one normal type (type 0) in the experiments to verify the superiority of our proposed method. The novel-class shot is two. Namely, two novel classes are added for training and testing for each incremental session, and thus there are five times new class increment.
As is shown in Table \ref{datasets}, for the normal class both the two imbalanced fault diagnosis experiments use 500 exemplars for training and 800 exemplars for testing. For the fault classes, we randomly set the numbers 48 and 30 of exemplars in each fault class for training under the two imbalanced fault diagnosis situations, respectively.


\subsubsection{Implementation Details}
In the representation learning stage of SCLIFD, we design a multi-layer perception (MLP). The encoder of the MLP has three layers and is used as the feature extractor network, with 52, 20, and 10 neurons in each layer respectively.
A fully connected layer is designed after the encoder, whose first layer has 10 neurons. And the neuron number of the second layer incrementally increases with the number of faults increasing. The maximum training epoch is set to 500. What's more, we train the network using Adam optimizer with mini-batches of 64 samples, and a weight decay of 0.00001. The initial learning rate is 0.01, which decays by gamma 0.2 once the epoch number reaches one of the milestones 200 and 400. The MLP is trained by the loss explained in section \ref{sec:SCLKD}. Additionally, we design a memory buffer of 
total size $K=100$ number of exemplars in the whole process of fault diagnosis. Finally, the models are assessed at each session using a joint set of testing sets of the classes encountered so far. PyTorch and scikit-learn are used in the whole implementation process.

\subsubsection{Comparative Results}
\begin{table*}[htbp]
\centering
\caption{Experiment Result Comparisons of Different Methods on TEP Dataset}
\label{TE-baseline-table}
\begin{threeparttable}          

\begin{tabular}{@{}cccccccccccc@{}}
\toprule
\multirow{2}{*}{Diagnosis Mode} & \multirow{2}{*}{Method} & \multicolumn{10}{c}{Accuracy and Relative Improvement in All Sessions(\%)}                                                      \\ \cmidrule(l){3-12} 
                                &                         & 1              & Impro. & 2              & Impro. & 3              & Impro. & 4              & Impro. & 5              & Impro. \\ \midrule
\multirow{7}{*}{Imbalanced case 1}     & LwF.MC                  & 99.49          & -0.11  & 28.57          & 47.71  & 19.35          & 73.25  & 12.18          & 70.73  & 9.8            & 71.62  \\
                                & Finetuning              & 99.09          & 0.29   & 46.99          & 29.29  & 32.16          & 60.44  & 18.63          & 64.28  & 17.73          & 63.69  \\
                                & iCaRL                   & 99.26          & 0.12   & 71.99          & 4.29   & 68.23          & 24.37  & 53.75          & 29.16  & 48.27          & 33.15  \\
                                & EEIL                    & 58.12          & 41.26  & 33.24          & 43.04  & 17.67          & 74.93  & 14.87          & 68.04  & 11.24          & 70.18  \\
                                & BiC                     & 59.55          & 39.83  & 32.57          & 43.71  & 25.35          & 67.25  & 19.63          & 63.28  & 17.76          & 63.66  \\
                                & \textbf{SCLIFD(Ours)}   & \textbf{99.38} & /      & \textbf{76.28} & /      & \textbf{92.6}  & /      & \textbf{82.91} & /      & \textbf{81.42} & /      \\
                                & \textit{Upper Bound}    & \textit{99.43} & /      & \textit{82.62} & /      & \textit{97.06} & /      & \textit{83.98} & /      & \textit{83.79} & /      \\ \midrule
\multirow{7}{*}{Imbalanced case 2}    & LwF.MC                  & 99.43          & -0.05  & 51.99          & 27.92  & 42.9           & 51.70  & 32.5           & 49.24  & 12.94          & 61.47  \\
                                & Finetuning              & 97.56          & 1.82   & 47.77          & 32.14  & 32.1           & 62.50  & 13.4           & 68.34  & 15.59          & 58.82  \\
                                & iCaRL                   & 99.26          & 0.12   & 72.35          & 7.56   & 63.51          & 31.09  & 48.26          & 33.48  & 38.57          & 35.84  \\
                                & EEIL                    & 57.81          & 41.57  & 27.14          & 52.77  & 13.38          & 81.22  & 10.26          & 71.48  & 8.14           & 66.27  \\
                                & BiC                     & 52.66          & 46.72  & 27.93          & 51.98  & 18.67          & 75.93  & 13.65          & 68.09  & 10.46          & 63.95  \\
                                & \textbf{SCLIFD(Ours)}   & \textbf{99.38} & /      & \textbf{79.91} & /      & \textbf{94.6}  & /      & \textbf{81.74} & /      & \textbf{74.41} & /      \\
                                & \textit{Upper Bound}    & \textit{99.32} & /      & \textit{95.3}  & /      & \textit{97.66} & /      & \textit{82.5}  & /      & \textit{77.45} & /      \\ \bottomrule
\end{tabular}
  \begin{tablenotes}    
        \footnotesize               
        \item[1] Impro. denotes Improvement. It represents the accuracy improvement of SCLIFD from the compared methods.
    \end{tablenotes}            
\end{threeparttable}          

\end{table*}

\begin{table*}[htbp]
\centering
\caption{Ablation Results on TEP Dataset}
\label{tab:TE-ablation}
\begin{threeparttable}          

\begin{tabular}{@{}cccccccccccccc@{}}
\toprule
\multirow{2}{*}{Diagnosis Mode} & \multicolumn{3}{c}{Components}             & \multicolumn{10}{c}{Accuracy and Relative Improvement in All Sessions(\%)}                                                                                             \\ \cmidrule(l){2-14} 
                                & SCL          & AdaHerding   & COS          & 1              & Impro.        & 2              & Impro.         & 3              & Impro.         & 4              & Impro.         & 5              & Impro.         \\ \midrule
\multirow{6}{*}{Imbalanced case 1}     & $\times$     & $\times$     & $\times$     & 99.26          & /             & 71.99          & /              & 68.23          & /              & 53.75          & /              & 48.27          & /              \\ \cmidrule(l){2-14} 
                                & $\checkmark$ & $\times$     & $\times$     & 99.38          & 0.12          & 73.21          & 1.22           & 83.41          & 15.18          & 77.56          & 23.81          & 76.23          & 27.96          \\     
                                & $\times$     & $\checkmark$ & $\times$     & 99.2           & -0.06         & 74.23          & 2.24           & 78.49          & 10.26          & 59.39          & 5.64           & 53.37          & 5.10           \\
                                & $\times$     & $\times$     & $\checkmark$& 99.26          & 0.00          & 73.2           & 1.21           & 74.81          & 6.58           & 56.36          & 2.61           & 51.13          & 2.86           \\\cmidrule(l){2-14} 
                                & $\checkmark$ & $\checkmark$ & $\times$     & 99.03          & -0.23         & 75.74          & 3.75           & 87.84          & 19.61          & 80.82          & 27.07          & 79.72          & 31.45          \\
                                & $\checkmark$ & $\checkmark$ & $\checkmark$ & \textbf{99.38} & \textbf{0.12} & \textbf{76.28} & \textbf{4.29}  & \textbf{92.6}  & \textbf{24.37} & \textbf{82.91} & \textbf{29.16} & \textbf{81.42} & \textbf{33.15} \\ \midrule
\multirow{6}{*}{Imbalanced case 2}    & $\times$     & $\times$     & $\times$     & 99.26          & /             & 72.35          & /              & 63.51          & /              & 48.26          & /              & 38.57          & /              \\\cmidrule(l){2-14} 
                                & $\checkmark$ & $\times$     & $\times$     & 99.2           & -0.06         & 75.03          & 2.68           & 72.58          & 9.07           & 66.72          & 18.46          & 56.45          & 17.88          \\
                                & $\times$     & $\checkmark$ & $\times$     & 99.26          & 0.00          & 74.82          & 2.47           & 68.52          & 5.01           & 54.2           & 5.94           & 42.45          & 3.88           \\
                                & $\times$     & $\times$     & $\checkmark$ & 99.2           & -0.06         & 73.88          & 1.53           & 68.29          & 4.78           & 50.73          & 2.47           & 41.06          & 2.49           \\\cmidrule(l){2-14} 
                                & $\checkmark$ & $\checkmark$ & $\times$     & 99.09          & -0.17         & 75.54          & 3.19           & 90.42          & 26.91          & 77.85          & 29.59          & 71.59          & 33.02          \\
                                & $\checkmark$ & $\checkmark$ & $\checkmark$ & \textbf{99.38} & \textbf{0.12} & \textbf{79.91} & \textbf{7.56}  & \textbf{94.6}  & \textbf{31.09} & \textbf{81.74} & \textbf{33.48} & \textbf{74.41} & \textbf{35.84} \\\bottomrule
\end{tabular}
  \begin{tablenotes}    
        \footnotesize               
        \item[1] Impro. denotes Improvement. It represents the accuracy improvement of each ablation method from the baseline iCaRL.
    \end{tablenotes}            

\end{threeparttable}          

\end{table*}

\begin{figure}[htbp]
    \centering
    \includegraphics[width=0.48\textwidth]{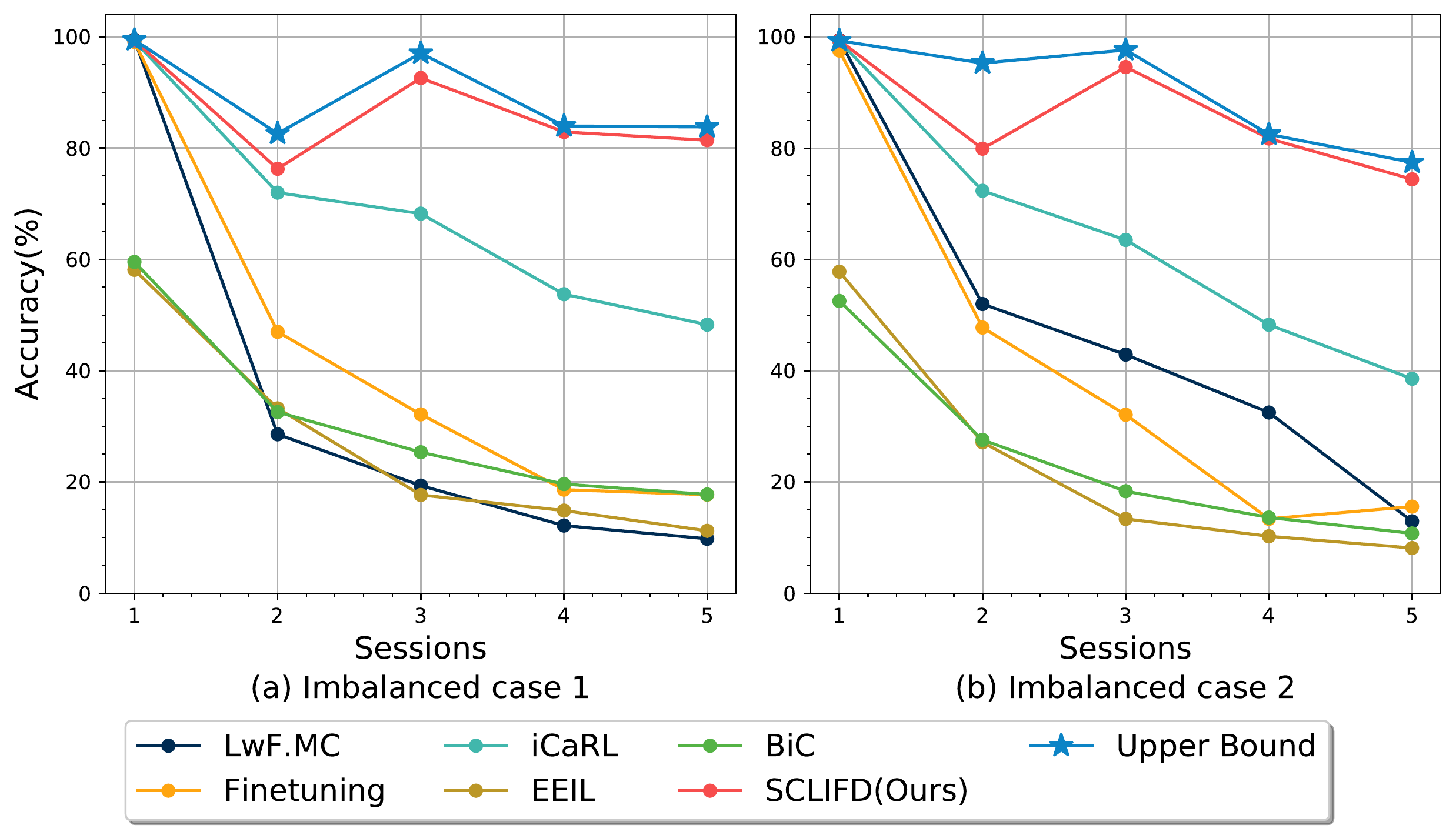}
    \caption{Experiment result comparisons of different methods on TEP Dataset: (a) imbalanced case 1 (b) imbalanced case 2}
    \label{fig:TE-baseline}
\end{figure}

\begin{figure*}[t]
    \centering
    \includegraphics[width=\textwidth]{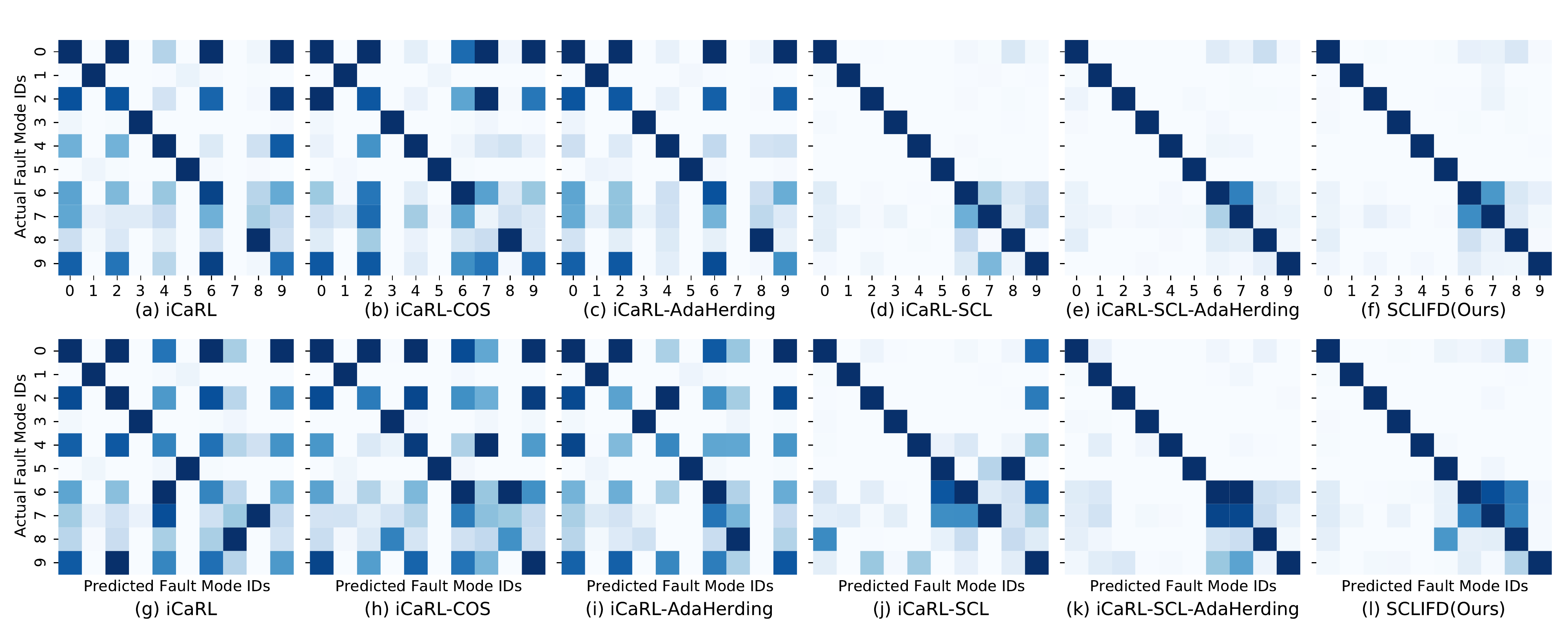}
    \vspace*{-8mm}
    \caption{The confusion matrices of ablation experiment results on \textbf{TEP dataset}(with entries scaled to range from 0 to 1 for better visibility). Fig.s from (a) to (f) are confusion matrices of \textbf{Imbalanced fault diagnosis case 1}. Fig.s from (g) to (l) are confusion matrices of \textbf{Imbalanced fault diagnosis case 2}. The darker the color is, the larger the number is.}
    \label{fig:TE_ablation_CM}
\end{figure*}

In this section, we report the comparative results of the methods under the circumstances of the two imbalanced fault diagnosis situations on the TEP dataset. Fig. \ref{fig:TE-baseline} shows the accuracy trends of all the different methods over all incremental sessions. Table \ref{TE-baseline-table} shows the test accuracies and relative improvement of our SCLIFD compared with state-of-the-art methods. The relative improvement is the difference between the accuracy of SCLIFD and other methods.
We summarize the experimental results as follows:
\begin{itemize}
    \item  For each encountered session under fault diagnosis of limited fault data, our SCLIFD always outperforms other state-of-the-art methods and is also the closest to the upper bound. 
    \item As incremental learning proceeds, the gap in the accuracy of SCLIFD and other methods is also continuously widening. This demonstrates that with incremental learning progresses, the superiority of SCLIFD becomes more apparent, exhibiting its ability to constantly learn longer sequences of new classes. 
    \item SCLIFD achieves the final accuracies of 81.42\% and 74.41\% under the two imbalanced fault diagnosis cases respectively, while the second best one (\textit{i}.\textit{e}., iCaRL) achieves the accuracies of 48.27\% and 38.57\% respectively. This demonstrates that other methods deteriorate the test accuracies drastically due to \textit{catastrophic forgetting}, while our SCLIFD effectively alleviates \textit{catastrophic forgetting}, outperforming iCaRL by up to 33.15\% and 35.84\%, and other methods even more.
    \item The outstanding performance of our SCLIFD may be on account of the following reasons. Primarily, SCL enormously enhances the representation learning ability of SCLIFD.
    Moreover, AdaHerding always assists SCLIFD to reserve the more complicated exemplars to revise at each incremental stage. Therefore, SCLIFD has a more significant ability to keep the knowledge learned from the old classes. Additionally, the cosine classifier mitigates the classifications biased towards the novel classes and the confusion among old classes due to the adverse effects of the data imbalance. Finally, the positive effects of the three components accumulate with the incremental learning proceeds. Accordingly, SCLIFD has distinguished performance on the imbalanced and long-tailed fault diagnosis.
\end{itemize}

\subsubsection{Ablation Study}
\label{TE-ablation}
\begin{figure}[htbp]
    \centering
    \includegraphics[width=0.48\textwidth]{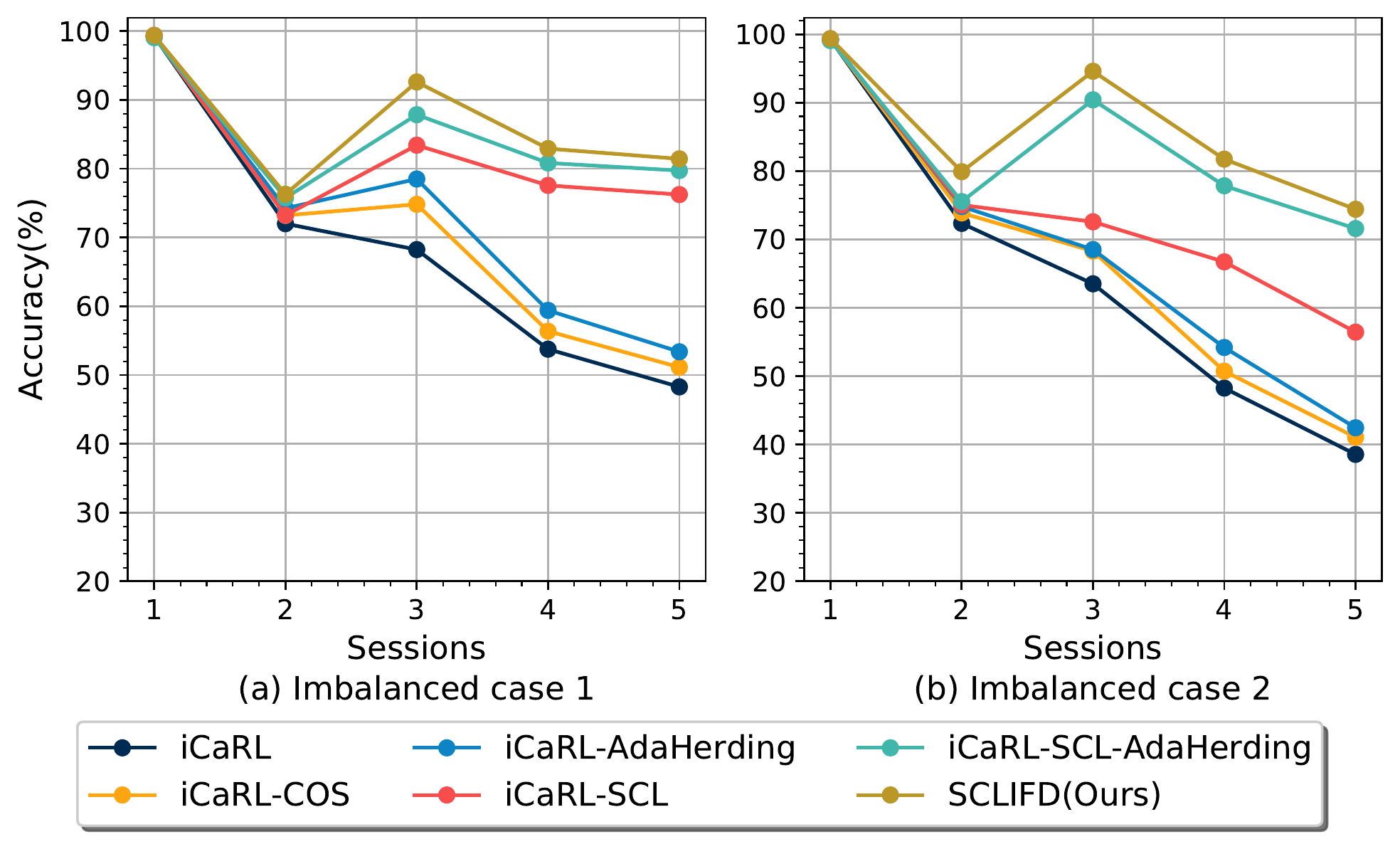}
    \caption{Ablation experiment results on the TEP Dataset: (a) imbalanced case 1 (b) imbalanced case 2}
    \label{fig:TE-ablation}
\end{figure}

Our approach mainly consists of three components, which are SCL, AdaHerding, and cosine classifier (COS). The results of the following intermediate models under the two imbalanced cases are provided to analyze the effect of individual components: iCaRL, iCaRL-SCL, iCaRL-AdaHerding, iCaRL-COS, iCaRL-SCL-AdaHerding, SCLIFD. Notably, the ablation methods adopt the herding and NME classifier in default. Namely, the method without using AdaHerding adopts herding. The method without using the cosine classifier adopts the NME classifier. Moreover, the methods with the SCL module mean it adopts both distillation loss and SCL loss. If not, it uses distillation loss only.

As is illustrated in Fig. \ref{fig:TE-ablation} and Table \ref{tab:TE-ablation}, the performance of the methods increases considerably when added SCL module, increases clearly when added AdaHerding or COS module. For instance, in imbalanced case 1, the accuracy increases 27.96\%, 5.10\%, and 2.86\% in iCaRL-SCL, iCaRL-AdaHerding, and iCaRL-COS respectively compared to iCaRL in the incremental session 5. The relative improvement is the difference between the accuracy of each ablation method and pure iCaRL.

Fig. \ref{fig:TE_ablation_CM} shows the confusion matrix of different ablation methods, providing further insight into the behaviors of the methods. It clearly shows that the predictions of the methods with SCL are much more on the diagonal entries (\textit{i}.\textit{e}. correction predictions) than those without SCL. 
The phenomenon demonstrates that SCL plays a prominent role in the outstanding performance of our method in the two imbalanced situations. Besides, AdaHerding and the COS module also contribute to the performance to a large extent.

\begin{table*}[htbp]
\label{tab:MFF-baseline}

\centering
\caption{Experiment result comparisons of different methods on MFF Dataset}
\begin{tabular}{@{}cccccccccccc@{}}
\toprule
\multirow{2}{*}{Diagnosis Mode} & \multirow{2}{*}{Method} & \multicolumn{10}{c}{Accuracy and Relative improvement in all Sessions(\%)}                                                      \\ \cmidrule(l){3-12} 
                                &                         & 1              & Impro. & 2              & Impro. & 3              & Impro. & 4              & Impro. & 5              & Impro. \\ \midrule
\multirow{7}{*}{Imbalanced}     & LwF.MC                  & 100            & 0.00   & 80.68          & 18.98  & 43.91          & 50.86  & 38.3           & 50.57  & 28.85          & 59.32  \\
                                & Finetuning              & 100            & 0.00   & 45.45          & 54.21  & 31.25          & 63.52  & 23.81          & 65.06  & 19.23          & 68.94  \\
                                & iCaRL                   & 100            & 0.00   & 74.15          & 25.51  & 75.02          & 19.75  & 55.68          & 33.19  & 57.16          & 31.01  \\
                                & EEIL                    & 78.43          & 21.57  & 43.34          & 56.32  & 33.87          & 60.90  & 25.93          & 62.94  & 22.19          & 65.98  \\
                                & BiC                     & 100            & 0.00   & 48.45          & 51.21  & 34.25          & 60.52  & 27.81          & 61.06  & 19.23          & 68.94  \\
                                & \textbf{SCLIFD(Ours)}   & \textbf{100}   & /      & \textbf{99.66} & /      & \textbf{94.77} & /      & \textbf{88.87} & /      & \textbf{88.17} & /      \\
                                & \textit{Upper Bound}    & \textit{100}   & /      & \textit{99.77} & /      & \textit{99.53} & /      & \textit{98.27} & /      & \textit{90.13} & /      \\ \midrule
\multirow{7}{*}{Long-Tailed}    & LwF.MC                  & 100            & 0.00   & 58.47          & 29.71  & 40.2           & 38.32  & 52.38          & 25.60  & 42.31          & 37.50  \\
                                 & Finetuning              & 100            & 0.00   & 45.45          & 42.73  & 31.25          & 47.27  & 23.81          & 54.17  & 19.23          & 60.58  \\
                                & iCaRL                   & 100            & 0.00   & 58.86          & 29.32  & 59.1           & 19.42  & 63.42          & 14.56  & 44.04          & 35.77  \\
                                & EEIL                    & 59.92          & 40.08  & 50.12          & 38.06  & 31.62          & 46.90  & 24.56          & 53.42  & 21.25          & 58.56  \\
                                & BiC                     & 100            & 0.00   & 41.74          & 46.44  & 30.10          & 48.42  & 21.53          & 56.45  & 17.12          & 62.69  \\
                                & \textbf{SCLIFD(Ours)}   & \textbf{100}   & /      & \textbf{88.18} & /      & \textbf{78.52} & /      & \textbf{77.98} & /      & \textbf{79.81} & /      \\
                                & \textit{Upper Bound}    & \textit{100}   & /      & \textit{90.17} & /      & \textit{88.66} & /      & \textit{85.88} & /      & \textit{84.45} & /      \\ \bottomrule
\end{tabular}
\end{table*}

\begin{table*}[htbp]
\centering
\caption{ABLATION RESULTS ON MFF DATASET}
\label{tab:MFF-ablation}
\begin{tabular}{@{}cccccccccccccc@{}}
\toprule
\multirow{2}{*}{Diagnosis Mode} & \multicolumn{3}{c}{Components}             & \multicolumn{10}{c}{Accuracy and Relative improvement in all Sessions(\%)}                                                                                             \\ \cmidrule(l){2-14} 
                                & SCL          & AdaHerding   & COS          & 1              & Impro.        & 2              & Impro.         & 3              & Impro.         & 4              & Impro.         & 5              & Impro.         \\ \midrule
\multirow{6}{*}{Imbalanced}     & $\times$     & $\times$     & $\times$     & 100            & /             & 74.15          & /              & 75.02          & /              & 55.68          & /              & 57.16          & /              \\\cmidrule(l){2-14} 
                                & $\checkmark$ & $\times$     & $\times$     & 100            & 0.00          & 81.76          & 7.61           & 82.07          & 7.05           & 79.52          & 23.84          & 79.02          & 21.86          \\
                                & $\times$     & $\checkmark$ & $\times$     & 100            & 0.00          & 83.13          & 8.98           & 87.3           & 12.28          & 69.38          & 13.70          & 63.58          & 6.42           \\
                                & $\times$     & $\times$     & $\checkmark$ & 100            & 0.00          & 80.06          & 5.91           & 84.3           & 9.28           & 78.13          & 22.45          & 76.85          & 19.69          \\     \cmidrule(l){2-14} 
                                & $\checkmark$ & $\checkmark$ & $\times$     & 100            & 0.00          & 87.63          & 13.48          & 89.58          & 14.56          & 81.99          & 26.31          & 82.24          & 25.08          \\
                                & $\checkmark$ & $\checkmark$ & $\checkmark$ & \textbf{100}   & \textbf{0.00} & \textbf{99.66} & \textbf{25.51} & \textbf{94.77} & \textbf{19.75} & \textbf{88.87} & \textbf{33.19} & \textbf{88.17} & \textbf{31.01} \\\midrule
\multirow{6}{*}{Long-Tailed}    & $\times$     & $\times$     & $\times$     & 100            & /             & 58.86          & /              & 59.1           & /              & 63.42          & /              & 44.04          & /              \\\cmidrule(l){2-14} 
                                & $\checkmark$ & $\times$     & $\times$     & 100            & 0.00          & 75.72          & 16.86          & 70.16          & 11.06          & 68.57          & 5.15           & 72.95          & 28.91          \\
                                & $\times$     & $\checkmark$ & $\times$     & 100            & 0.00          & 65             & 6.14           & 63.01          & 3.91           & 66.72          & 3.30           & 61.03          & 16.99          \\
                                & $\times$     & $\times$     & $\checkmark$     & 100            & 0.00          & 63.69          & 4.83           & 64.92          & 5.82           & 71.61          & 8.19           & 64.13          & 20.09          \\\cmidrule(l){2-14} 
                                & $\checkmark$ & $\checkmark$ & $\times$     & 100            & 0.00          & 76.31          & 17.45          & 71.45          & 12.35          & 75.65          & 12.23          & 78.46          & 34.42          \\
                                & $\checkmark$ & $\checkmark$ & $\checkmark$ & \textbf{100}   & \textbf{0.00} & \textbf{88.18} & \textbf{29.32} & \textbf{78.52} & \textbf{19.42} & \textbf{77.98} & \textbf{14.56} & \textbf{79.81} & \textbf{35.77} \\ \bottomrule
\end{tabular}
\end{table*}

\subsection{Practical Case Study on MFF}
In the above section, we have conducted numerous experiments on the simulated TEP process. In this section, to further verify the effectiveness of our proposed method, we extend the experiments on the realistic Multiphase Flow Facility (MFF) dataset, which is collected  in the actual industrial scenarios. What's more, apart from the imbalanced fault diagnosis, we also evaluate the long-tailed fault diagnosis of the methods.
\subsubsection{Dataset Description}
The benchmark dataset is conducted on a multiphase flow facility, where water and air are combined and then separated as they move across the horizontal part. A variety of different testing scenarios are used to generate data for both normal and problematic states. Multiple pressure, flow rate, temperature, and density sensors are installed throughout the facility \cite{ruiz2015statistical}. 

The dataset is comprised of 6 types of faults, from which we randomly pick faults 1, 2, 3, and 4 in the experiments. Class 0 is the normal class. As is shown in Table \ref{datasets}, one novel class is added for each incremental session. Hence there are 5 incremental sessions in this part. For the normal class, both the imbalanced and long-tailed fault diagnosis experiments use 500 exemplars for training and 800 exemplars for testing. For the fault classes, we set the numbers 30 and 20 of exemplars in each fault class for training under the imbalanced and long-tailed fault diagnosis situations \cite{Peng_Lu_Tao_Ma_Zhang_Wang_Zhang_2022}, respectively.

\subsubsection{Implementation Details}
Generally, the implementation of this section is the same as that of the evaluation of the TEP dataset, except for the encoder of the MLP and the size of the memory buffer. 
The encoder of the MLP has 24, 12, and 10 neurons in each layer respectively. 
Additionally, the memory buffer size $K$ is designed to be 40 in this part.

\subsubsection{Comparative Results}
\begin{figure}[htbp]
    \centering
    \includegraphics[width=0.48\textwidth]{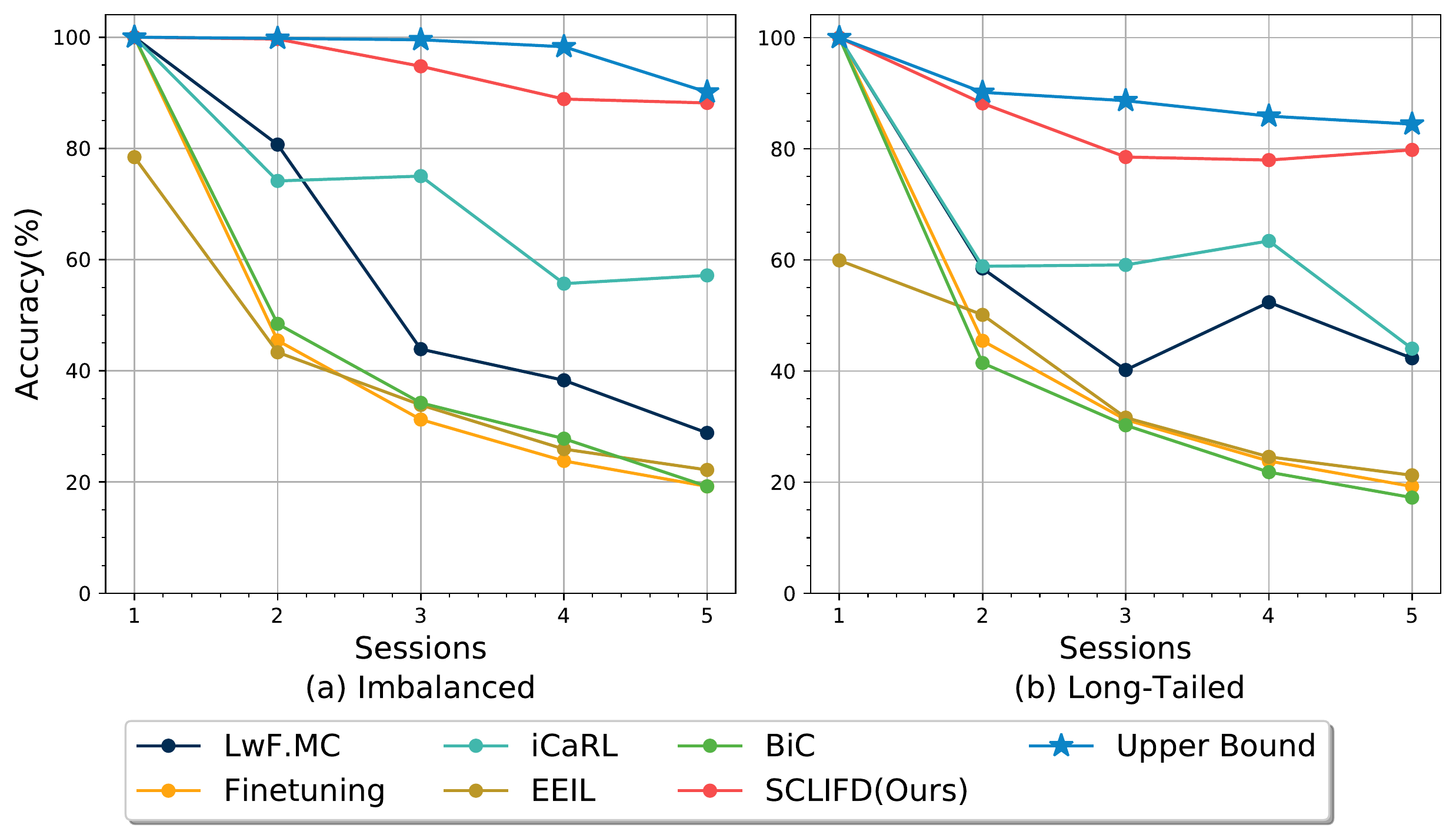}
    \caption{Experiment result comparisons of different methods on MFF Dataset: (a) imbalanced fault diagnosis (b) long-tailed fault diagnosis.}
    \label{fig:MFF-baseline}
\end{figure}

In this section, we elaborate on the comparative results of the different methods under the imbalanced and long-tailed fault diagnosis situations on the MFF dataset. Fig. \ref{fig:MFF-baseline} shows the accuracy trends over all incremental sessions. Table \ref{tab:MFF-baseline} shows the test accuracies and relative improvement of our SCLIFD compared with other methods. 
SCLIFD achieves the final accuracies of 88.17\% and 79.81\% under the imbalanced and long-tailed fault diagnosis cases respectively, while the second best one (\textit{i}.\textit{e}. iCaRL) achieves the accuracies of 57.16\% and 44.04\% respectively. Our SCLIFD outperforms iCaRL by up to 33.15\% and 35.84\%, and other methods even more.

For each encountered session our SCLIFD always outperforms other state-of-the-art methods by a large margin and is also the closest to the upper bound. This further proves the superiority of our SCLIFD method.


\subsubsection{Ablation Study}
\begin{figure}[htbp]
    \centering
    \includegraphics[width=0.48\textwidth]{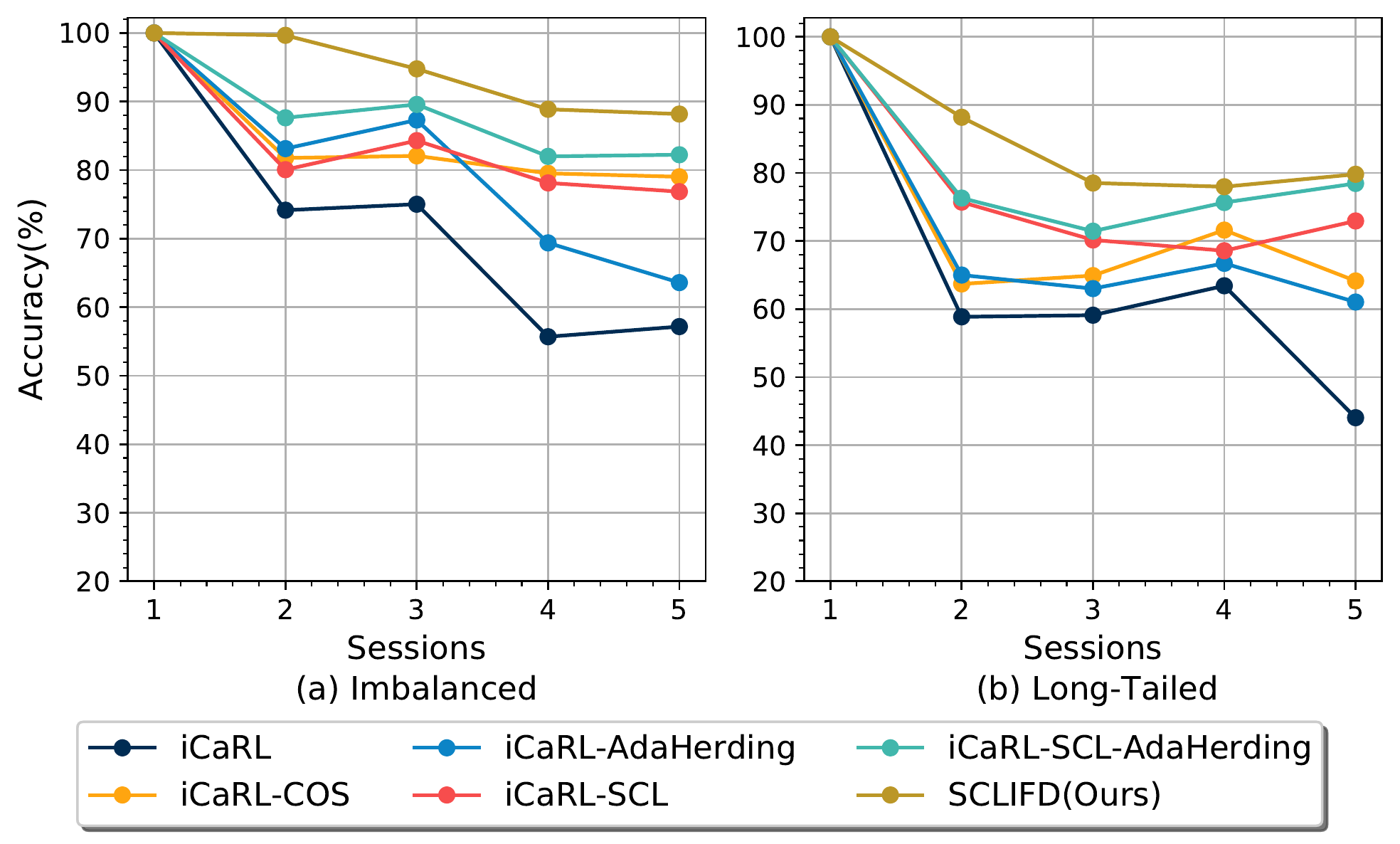}
    \caption{Ablation experiment results on MFF Dataset: (a) imbalanced fault diagnosis (b) long-tailed fault diagnosis}
    \label{fig:MFF-ablation}
\end{figure}

In this section, we still mainly focus on the three components, which are SCL, AdaHerding, and COS classifier. The intermediate models under the imbalanced and long-tailed situations are the same as those in Table \ref{tab:TE-ablation}. As is illustrated in Fig. \ref{fig:MFF-ablation} and Table \ref{tab:MFF-ablation}, for instance in imbalanced fault diagnosis, the accuracy increases 21.86\%, 6.42\%, 19.69\% in iCaRL-SCL, iCaRL-AdaHerding, and iCaRL-COS respectively compared to iCaRL in the incremental session 5. This demonstrates that SCL, AdaHerding, and the COS modules all contribute to the performance improvement of SCLIFD.

\section{Conclusion}
\label{conclusion}
In this paper, we propose the SCLIFD framework for fault diagnosis under inadequate fault data in the class incremental learning setting. SCLIFD utilizes supervised contrastive knowledge
distillation to boost its representation learning capability and alleviate catastrophic forgetting. Furthermore, we propose the AdaHerding method to select exemplars by giving priority to their learning difficulty when constructing the exemplar set. Additionally, we apply the cosine classifier at the classification stage to mitigate the negative impact of class imbalance.
We evaluate the SCLIFD method in two common settings: imbalanced fault diagnosis and long-tailed fault diagnosis and conduct extensive experiments on the simulated TE process dataset and the real-world MFF process dataset. As illustrated in the experiments, the SCLIFD method has achieved the best performances in different settings, compared to other state-of-the-art methods. This demonstrates that SCLIFD provides a novel perspective to tackle the challenging class incremental learning in fault diagnosis under inadequate fault data. 

In future work, we will pay more attention to the interpretability of common fault diagnosis, especially the class incremental fault diagnosis studied in the paper, which may be conducive to further enhancing the performance of the proposed method. Moreover, we would consider the scenarios where previous data cannot be stored for certain reasons, such as privacy considerations.


%


\ifCLASSOPTIONcaptionsoff
  \newpage
\fi



%
\footnotesize{
\bibliographystyle{IEEEtranN}

\bibliography{ref.bib}

\begin{thebibliography}{27}
\providecommand{\natexlab}[1]{#1}
\providecommand{\url}[1]{#1}
\csname url@samestyle\endcsname
\providecommand{\newblock}{\relax}
\providecommand{\bibinfo}[2]{#2}
\providecommand{\BIBentrySTDinterwordspacing}{\spaceskip=0pt\relax}
\providecommand{\BIBentryALTinterwordstretchfactor}{4}
\providecommand{\BIBentryALTinterwordspacing}{\spaceskip=\fontdimen2\font plus
\BIBentryALTinterwordstretchfactor\fontdimen3\font minus
  \fontdimen4\font\relax}
\providecommand{\BIBforeignlanguage}[2]{{%
\expandafter\ifx\csname l@#1\endcsname\relax
\typeout{** WARNING: IEEEtranN.bst: No hyphenation pattern has been}%
\typeout{** loaded for the language `#1'. Using the pattern for}%
\typeout{** the default language instead.}%
\else
\language=\csname l@#1\endcsname
\fi
#2}}
\providecommand{\BIBdecl}{\relax}
\BIBdecl

\bibitem[Zou et~al.(2018)Zou, Xia, and Li]{Zou_Xia_Li_2018}
W.~Zou, Y.~Xia, and H.~Li, ``Fault diagnosis of tennessee-eastman process using
  orthogonal incremental extreme learning machine based on driving amount,''
  \emph{IEEE Transactions on Cybernetics}, vol.~48, no.~12, p. 3403–3410, Dec
  2018.

\bibitem[Chen et~al.(2022{\natexlab{a}})Chen, Chen, Fan, Peng, and
  Yang]{Chen_Chen_Fan_Peng_Yang_2022}
Z.~Chen, W.~Chen, X.~Fan, T.~Peng, and C.~Yang, ``Jitl-mbn: A real-time
  causality representation learning for sensor fault diagnosis of traction
  drive system in high-speed trains,'' \emph{IEEE Transactions on Neural
  Networks and Learning Systems}, p. 1–12, 2022.

\bibitem[Wang et~al.(2021)Wang, Liu, Lin, Chen, Li, Hu, and
  Chen]{Wang_Liu_Lin_Chen_Li_Hu_Chen_2021}
Y.~Wang, R.~Liu, D.~Lin, D.~Chen, P.~Li, Q.~Hu, and C.~L.~P. Chen,
  ``Coarse-to-fine: Progressive knowledge transfer-based multitask
  convolutional neural network for intelligent large-scale fault diagnosis,''
  \emph{IEEE Transactions on Neural Networks and Learning Systems}, p. 1–14,
  2021.

\bibitem[Peng et~al.(2022)Peng, Lu, Tao, Ma, Zhang, Wang, and
  Zhang]{Peng_Lu_Tao_Ma_Zhang_Wang_Zhang_2022}
P.~Peng, J.~Lu, S.~Tao, K.~Ma, Y.~Zhang, H.~Wang, and H.~Zhang, ``Progressively
  balanced supervised contrastive representation learning for long-tailed fault
  diagnosis,'' \emph{IEEE Transactions on Instrumentation and Measurement},
  vol.~71, p. 1–12, 2022.

\bibitem[Chen et~al.(2022{\natexlab{b}})Chen, Chen, Feng, Liu, Zhang, Zhang,
  and Xiao]{Chen_Chen_Feng_Liu_Zhang_Zhang_Xiao_2022}
Z.~Chen, J.~Chen, Y.~Feng, S.~Liu, T.~Zhang, K.~Zhang, and W.~Xiao,
  ``\BIBforeignlanguage{en}{Imbalance fault diagnosis under long-tailed
  distribution: Challenges, solutions and prospects},''
  \emph{\BIBforeignlanguage{en}{Knowledge-Based Systems}}, vol. 258, p. 110008,
  Dec 2022.

\bibitem[Khoshgoftaar and Gao(2009)]{Khoshgoftaar_Gao_2009a}
T.~M. Khoshgoftaar and K.~Gao, ``Feature selection with imbalanced data for
  software defect prediction,'' in \emph{2009 International Conference on
  Machine Learning and Applications}, Dec 2009, p. 235–240.

\bibitem[Liu et~al.(2017)Liu, Li, and Zio]{Liu_Li_Zio_2017}
J.~Liu, Y.-F. Li, and E.~Zio, ``\BIBforeignlanguage{en}{A svm framework for
  fault detection of the braking system in a high speed train},''
  \emph{\BIBforeignlanguage{en}{Mechanical Systems and Signal Processing}},
  vol.~87, p. 401–409, Mar 2017.

\bibitem[Goodfellow et~al.(2020)Goodfellow, Pouget-Abadie, Mirza, Xu,
  Warde-Farley, Ozair, Courville, and
  Bengio]{Goodfellow_Pouget-Abadie_Mirza_Xu_Warde-Farley_Ozair_Courville_Bengio_2020}
I.~Goodfellow, J.~Pouget-Abadie, M.~Mirza, B.~Xu, D.~Warde-Farley, S.~Ozair,
  A.~Courville, and Y.~Bengio, ``Generative adversarial networks,''
  \emph{Communications of the ACM}, vol.~63, no.~11, p. 139–144, Oct 2020.

\bibitem[Li et~al.(2021)Li, Jiang, Liu, Zhang, and
  Xu]{Li_Jiang_Liu_Zhang_Xu_2021}
X.~Li, H.~Jiang, S.~Liu, J.~Zhang, and J.~Xu, ``\BIBforeignlanguage{en}{A
  unified framework incorporating predictive generative denoising autoencoder
  and deep coral network for rolling bearing fault diagnosis with unbalanced
  data},'' \emph{\BIBforeignlanguage{en}{Measurement}}, vol. 178, p. 109345,
  Jun 2021.

\bibitem[Zhiyi et~al.(2020)Zhiyi, Haidong, Lin, Junsheng, and
  Yu]{Zhiyi_Haidong_Lin_Junsheng_Yu_2020}
H.~Zhiyi, S.~Haidong, J.~Lin, C.~Junsheng, and Y.~Yu,
  ``\BIBforeignlanguage{en}{Transfer fault diagnosis of bearing installed in
  different machines using enhanced deep auto-encoder},''
  \emph{\BIBforeignlanguage{en}{Measurement}}, vol. 152, p. 107393, Feb 2020.

\bibitem[Chu et~al.(2020)Chu, Bian, Liu, and Ling]{chu2020feature}
P.~Chu, X.~Bian, S.~Liu, and H.~Ling, ``Feature space augmentation for
  long-tailed data,'' in \emph{European Conference on Computer Vision}.\hskip
  1em plus 0.5em minus 0.4em\relax Springer, 2020, pp. 694--710.

\bibitem[McCloskey and Cohen(1989)]{McCloskey_Cohen_1989}
M.~McCloskey and N.~J. Cohen, \emph{\BIBforeignlanguage{en}{Catastrophic
  Interference in Connectionist Networks: The Sequential Learning
  Problem}}.\hskip 1em plus 0.5em minus 0.4em\relax Academic Press, Jan 1989,
  vol.~24, p. 109–165.

\bibitem[Picot et~al.(2018)Picot, Rivière, and Maussion]{Picot_Maussion_2018}
A.~Picot, J.~Rivière, and P.~Maussion, ``Bearing fault diagnosis based on the
  analysis of recursive pca projections,'' in \emph{2018 IEEE International
  Conference on Industrial Technology (ICIT)}, Feb 2018, p. 2093–2098.

\bibitem[Hell et~al.(2022)Hell, Pestana~de Aguiar, Soares, and
  Goliatt]{Hell_Pestana_Soares_Goliatt_2022}
M.~Hell, E.~Pestana~de Aguiar, N.~Soares, and L.~Goliatt,
  ``\BIBforeignlanguage{en}{A data-driven time-series fault prediction
  framework for dynamically evolving large-scale data streaming systems},''
  \emph{\BIBforeignlanguage{en}{International Journal of Fuzzy Systems}},
  vol.~24, no.~6, p. 2831–2844, Sep 2022.

\bibitem[Ren et~al.(2022)Ren, Liu, Wang, and Zhang]{Ren_Liu_Wang_Zhang_2022}
Y.~Ren, J.~Liu, Q.~Wang, and H.~Zhang, ``Hsell-net: A heterogeneous sample
  enhancement network with lifelong learning under industrial small samples,''
  \emph{IEEE Transactions on Cybernetics}, p. 1–13, 2022.

\bibitem[Tao et~al.(2020)Tao, Hong, Chang, Dong, Wei, and
  Gong]{Tao_Hong_Chang_Dong_Wei_Gong_2020}
X.~Tao, X.~Hong, X.~Chang, S.~Dong, X.~Wei, and Y.~Gong,
  ``\BIBforeignlanguage{en}{Few-shot class-incremental learning},'' in
  \emph{\BIBforeignlanguage{en}{2020 IEEE/CVF Conference on Computer Vision and
  Pattern Recognition (CVPR)}}.\hskip 1em plus 0.5em minus 0.4em\relax Seattle,
  WA, USA: IEEE, Jun 2020, p. 12180–12189.

\bibitem[Castro et~al.(2018)Castro, Marin-Jimenez, Guil, Schmid, and
  Alahari]{Castro_2018_ECCV}
F.~M. Castro, M.~J. Marin-Jimenez, N.~Guil, C.~Schmid, and K.~Alahari,
  ``End-to-end incremental learning,'' in \emph{Proceedings of the European
  Conference on Computer Vision (ECCV)}, September 2018.

\bibitem[Wu et~al.(2019)Wu, Chen, Wang, Ye, Liu, Guo, and Fu]{Wu_2019_CVPR}
Y.~Wu, Y.~Chen, L.~Wang, Y.~Ye, Z.~Liu, Y.~Guo, and Y.~Fu, ``Large scale
  incremental learning,'' in \emph{Proceedings of the IEEE/CVF Conference on
  Computer Vision and Pattern Recognition (CVPR)}, June 2019.

\bibitem[Welling(2009)]{welling2009herding}
M.~Welling, ``Herding dynamical weights to learn,'' in \emph{Proceedings of the
  26th Annual International Conference on Machine Learning}, 2009, pp.
  1121--1128.

\bibitem[Settles(2009)]{settles2009active}
B.~Settles, ``Active learning literature survey,'' 2009.

\bibitem[Hu et~al.(2021)Hu, Wu, Jian, Chen, and Yu]{Hu_Wu_Jian_Chen_Yu_2021}
W.~Hu, L.~Wu, M.~Jian, Y.~Chen, and H.~Yu, ``\BIBforeignlanguage{en}{Cosine
  metric supervised deep hashing with balanced similarity},''
  \emph{\BIBforeignlanguage{en}{Neurocomputing}}, vol. 448, p. 94–105, Aug
  2021.

\bibitem[Xu et~al.(2022)Xu, Yu, and Essaf]{Xu_Yu_Essaf_2022}
Q.~Xu, N.~Yu, and F.~Essaf, ``\BIBforeignlanguage{en}{Improved wafer map
  inspection using attention mechanism and cosine normalization},''
  \emph{\BIBforeignlanguage{en}{Machines}}, vol.~10, no.~22, p. 146, Feb 2022.

\bibitem[Koto(2014)]{Koto_2014}
F.~Koto, ``Smote-out, smote-cosine, and selected-smote: An enhancement strategy
  to handle imbalance in data level,'' in \emph{2014 International Conference
  on Advanced Computer Science and Information System}, Oct 2014, p. 280–284.

\bibitem[Khosla et~al.(2020)Khosla, Teterwak, Wang, Sarna, Tian, Isola,
  Maschinot, Liu, and
  Krishnan]{Khosla_Teterwak_Wang_Sarna_Tian_Isola_Maschinot_Liu_Krishnan_2020}
P.~Khosla, P.~Teterwak, C.~Wang, A.~Sarna, Y.~Tian, P.~Isola, A.~Maschinot,
  C.~Liu, and D.~Krishnan, ``Supervised contrastive learning,'' in
  \emph{Advances in Neural Information Processing Systems}, vol.~33.\hskip 1em
  plus 0.5em minus 0.4em\relax Curran Associates, Inc., 2020, p. 18661–18673.

\bibitem[Rebuffi et~al.(2017)Rebuffi, Kolesnikov, Sperl, and Lampert]{8100070}
S.-A. Rebuffi, A.~Kolesnikov, G.~Sperl, and C.~H. Lampert, ``icarl: Incremental
  classifier and representation learning,'' in \emph{2017 IEEE Conference on
  Computer Vision and Pattern Recognition (CVPR)}, 2017, pp. 5533--5542.

\bibitem[Li and Hoiem(2018)]{Li2018}
Z.~Li and D.~Hoiem, ``Learning without forgetting,'' \emph{IEEE Transactions on
  Pattern Analysis and Machine Intelligence}, vol.~40, pp. 2935--2947, 12 2018.

\bibitem[Ruiz-C{\'a}rcel et~al.(2015)Ruiz-C{\'a}rcel, Cao, Mba, Lao, and
  Samuel]{ruiz2015statistical}
C.~Ruiz-C{\'a}rcel, Y.~Cao, D.~Mba, L.~Lao, and R.~Samuel, ``Statistical
  process monitoring of a multiphase flow facility,'' \emph{Control Engineering
  Practice}, vol.~42, pp. 74--88, 2015.

\end{thebibliography}
}
\vspace{12pt}
\end{document}